\documentclass[review]{elsarticle}
 
\usepackage[T1]{fontenc}
\usepackage[utf8]{inputenc}
\usepackage{lmodern}
\usepackage[english]{babel}
\usepackage{enumitem}
\usepackage{amsmath}
\usepackage{amssymb}
\usepackage{graphicx}
\usepackage{wrapfig}
\usepackage{float}
\usepackage{xcolor}
\usepackage{algorithm}
\usepackage{algorithmic}
\usepackage{multirow}
\usepackage{bbm}
\usepackage{lineno}

\usepackage{fancyhdr}
\newenvironment{definition}[1][Definition]{\begin{trivlist}
\item[\hskip \labelsep {\bfseries #1}]}{\end{trivlist}}
\newenvironment{prop}[1][Proposition]{\begin{trivlist}
\item[\hskip \labelsep {\bfseries #1}]}{\end{trivlist}}
  
\begin{document}

\begin{frontmatter}
\title{Unsupervised Feature Selection Based on Space Filling Concept}
\author{Mohamed Laib and Mikhail Kanevski}
\address{Institute of Earth Surface Dynamics, Faculty of Geosciences and Environment, University of Lausanne, 1015 Lausanne, Switzerland. Email: Mohamed.Laib@unil.ch.}

\begin{abstract}

The paper deals with the adaptation of a new measure for the unsupervised feature selection problems. The proposed measure is based on space filling concept and is called the coverage measure. This measure was used for judging the quality of an experimental space filling design. In the present work, the coverage measure is adapted for selecting the smallest informative subset of variables by reducing redundancy in data. This paper proposes a simple analogy to apply this measure. It is implemented in a filter algorithm for unsupervised feature selection problems. 

The proposed filter algorithm is robust with high dimensional data and can be implemented without extra parameters. Further, it is tested with simulated data and real world case studies including environmental data and hyperspectral image. Finally, the results are evaluated by using random forest algorithm.

\end{abstract}

\begin{keyword}
{Unsupervised feature selection \sep Coverage measure  \sep Space filling \sep Random forest \sep Machine Learning }
\end{keyword}

\end{frontmatter}


\section{Introduction}
\label{intro}

In recent years, the techniques of collecting environmental data (such as: wind speed, permafrost, rainfall, pollution $\ldots$) have been improved. Moreover, environmental phenomena are mostly: non-linear, multivariate, and in many cases they are studied in high dimensional feature spaces \cite{KPT2009}. Usually, the input space is constructed by considering available information and expert knowledge. The empirically designed input feature space can gain rapidly a high dimension. In addition to the original features, there is always redundancy in the input data. In fact, the data points are not uniformly distributed in the experimental domain in which the data are embedded. In other word, the data space is not well filled or covered in the presence of redundancy. Consequently, the modelling of these data could take much time when introducing all features. Such problems are known as the curse of dimensionality.

To overcome this issue, feature selection (FS) algorithms play an important role in data driven modelling. Therefore, numerous methods and measures for FS have been proposed \cite{intro1,intro2}. The main purpose is to retain only features that bring new and relevant information by reducing the existing redundancy in data. This procedure helps to manage the curse of dimensionality. In fact, it improves the accuracy of modelling, speeds up the learning process, and offers a good interpretation of the results.

The literature of machine learning distinguishes two well-known techniques of FS, according to the availability of the output variable: supervised and unsupervised feature selection \cite{jjg1, jjg2, jango1, jango2}. These techniques try to find the smallest informative subset of features regarding to a defined measure or criterion.

Other methods are available, such as feature ranking \cite{rank1,rank2}, which consists in giving an order to features regarding their importance. Then, a learning process to choose how many features can be selected usually follows these methods.

Several measures and criteria are used for selecting the smallest subset of features: measures based on entropy \cite{entro1, entro2, entro3}, fractal dimension \cite{frac1}, intrinsic dimension \citep{jango1, jango2,intd1}, and also on distance \cite{dist1}.

In the unsupervised methods, the goal is mainly to carry out an exploratory analysis and to improve the discovering of hidden pattern. Therefore, the techniques of unsupervised feature selection (UFS) \cite{unsu1,unsu2,unsu3} do not require a prior information (output variable). They try to minimise existing redundancy, which leads to a reduction of dimensionality of data. Further, UFS techniques improve the understanding, the visualisation, and the interpretation of the results. In short, the dimensionality reduction consists in choosing a subset of features that contain new and relevant information about the data.

This paper is an adaptation of a new measure based on space filling concept, which is called the coverage measure. It was mainly used in experimental designs \cite{cov1,cov2}.
Moreover, the proposed measure was used also for the construction of spatial coverage designs in \cite{cov3}, which proposes its implementation in Splus. Other implementation for  the spatial coverage is available in the R library \textsf{spcosa} proposed in \cite{cov4}.
The \textsf{DiceDesign} R library proposes an implementation of this measure, in the context of space filling design \cite{cov5}.

The coverage measure is adapted here for the UFS problems. It can be implemented in all search techniques such as exhaustive search \cite{exau1} , sequential forward selection (SFS) \cite{sfs1}, and sequential backward selection \cite{sbf1} (SBS). In this work, it is considered with a SFS technique.

The proposed measure computes how well the space is covered by the data points. In fact, it quantifies the uniformity of points in a hypercube by comparing the repartition of points to a regular grid \cite{cov1}. The smallest coverage value means that the hypercube is well filled. Intuitively, the coverage measure gives zero value or near to zero if the data points are distributed as a regular grid, or near to be a regular grid, in the data space. 

The analogy is quite simple and clear, the selected features have to fill uniformly all the space in which the data are incorporated. In fact, the repartition of points expresses the information amount disseminated. Therefore, the smallest value of coverage means that the variables cover well the space in which they are embedded. Moreover, the selected variables should contain new and relevant information about the data. 

A filter algorithm is used to implement the coverage measure. It is applied on simulated dataset and on several well-known benchmark datasets used for feature selection purpose. In addition, real environmental data are used as well. 

Further, the algorithm is tested with different scenarios of noise injection and shuffling data. Then, the results are verified and evaluated with random forest algorithm \cite{rf1,rf2} by using a consistent methodology. 

The remainder of this paper is organized as follows. Section 2 explains the coverage measure and its use in experimental designs. Section 3 presents the implementation of this measure for the UFS problems and introduces the corresponding filter algorithm. In section 4, the measure is evaluated by several datasets. In the last section, the conclusion is given with future developments.

\section{Definition and basic notions}
\label{def}

Design and modelling of experiments have always been a fundamental  approach over the years. The experimenter has to propose and choose the suitable factor space (i.e. experimental domain) for the experiment under study.
The most important early step to check is the coverage or the uniformity of the proposed design. There are many ways to select the best design regarding several conditions and criteria \cite{fangbo}. 

Numerous space filling design have been proposed under some prior properties. They can be constructed by using algebraic methods: based on incomplete block resolvable design \cite{fangbo}, based on association schemes \cite{imane}. The construction algorithms were as well considered in \citep{pack1, pack2, pack3}. Other high quality designs, based on space filling concept, were proposed in  \cite{sf1,sf2,sf3}. Furthermore, different measures for choosing the best design have been given in \cite{meas1, meas2, meas3,damblin}. 

In the literature of sampling methods, one strategy is to generate randomly different designs. Then, a comparison is carried out using a defined measure to find the best design. Another approach can be on the extension of an existing design. The objective is to add more points in the sampling design by taking into account the prior defined measure.

Other strategies in choosing the best design is to adopt some optimality criterion, such as:
\begin{itemize}
\item The entropy criterion \cite{shannon,con_ent}, which has been widely used in coding theory and statistics. The Shannon entropy measures the amount of information contained in the distribution of a set of points. In \cite{con_ent} it is described as the classical idea of the information amount in an experiment. Moreover, it is proposed with a linear model (a simple Kriging model), and presents the corresponding maximum entropy designs.

\item The integrated mean squared error \citep{imse}, which is computationally more demanding and needs a powerful optimisation algorithm due to the large combinatorial design space. This criterion can be replaced by the maximum mean squared error involving a multidimensional optimisation \cite{mmse}.

\item Minimax or Maximin distance criteria, proposed in \cite{minmax}, which measure how well the experimental points are distributed through the experimental domain. A minimax distance minimises the maximum distance between points. Whereas the maximin distance maximises the minimum inter-site distance. A well-known maximin designs are the Placket-Burman designs where the number of points $n= 4m+1$ where $m$ is a positive integer  and presents the number of factors. 
\end{itemize}

Besides, several uniformity measures have been proposed in \cite{fangbo}. The most known is the discrepancy. Numerous kinds of discrepancies have been defined such as: the star discrepancy, the centred $L_2$-discrepancy, and wrap-around $L_2$-discrepancy. These uniformity criteria are based on the Kolmogorov-Smirnov test. In fact, it compares the design to a uniform distribution.

In addition to the discrepancy, the coverage measure was also proposed to quantify the uniformity. In contrast to the discrepancy, the coverage measure compares the proposed design to a regular grid. Furthermore, the coverage measure is more stable than the discrepancy in a high dimensional design. Therefore, it can be applied to high dimensional data.

\subsection{Coverage measure}
\label{2.1}
\begin{definition}

Let $X = \left\lbrace x^{1}, \ldots,  x^{n} \right\rbrace  \subset [0, 1]^{d}$  be a sequence of $n$ points of dimension $d$. The coverage measure is defined as follows:
\begin{equation}
\lambda = \frac{1}{\bar{\vartheta}}\left( \frac{1}{n}\sum\limits^{n}_{i=1} (\vartheta_{i}- \bar{\vartheta})^{2}    \right)^{\frac{1}{2}}
\end{equation}\label{equ_1} 
where:
$\vartheta_{i} = \min_k \: ( dist \left( x^{i},x^{k}  \right))$ is the minimal distance between $x^{i}$ and the other points of the sequence. And: $\bar{\vartheta} = \frac{1}{n} \sum\limits^{n}_{i=1} \vartheta_{i}$ is the mean of $\vartheta_{i}$; where  $dist$ is the euclidean distance.
\end{definition}

If the data points are distributed as a regular grid: $\vartheta_{1}=\vartheta_{2}=\ldots=\vartheta_{n}=\bar{\vartheta}$. Hence, $\lambda=0$

The quality coverage of points can be detected by using the minimum euclidean distance between the points. Further, it takes into account the dispersion of distances. In fact, the coverage measure $\lambda$ makes appear the coefficient of variation of the $\vartheta_{i}$, which is known as the relative standard deviation (the ratio of the standard deviation to the mean of $\vartheta_{i}$).

The smaller the value of $\lambda$ is, the smaller the distance between the points is. In this case, the design is near to be a regular grid. The best design should have the smallest coverage value $\lambda$.

\begin{figure}
\includegraphics[scale=0.11]{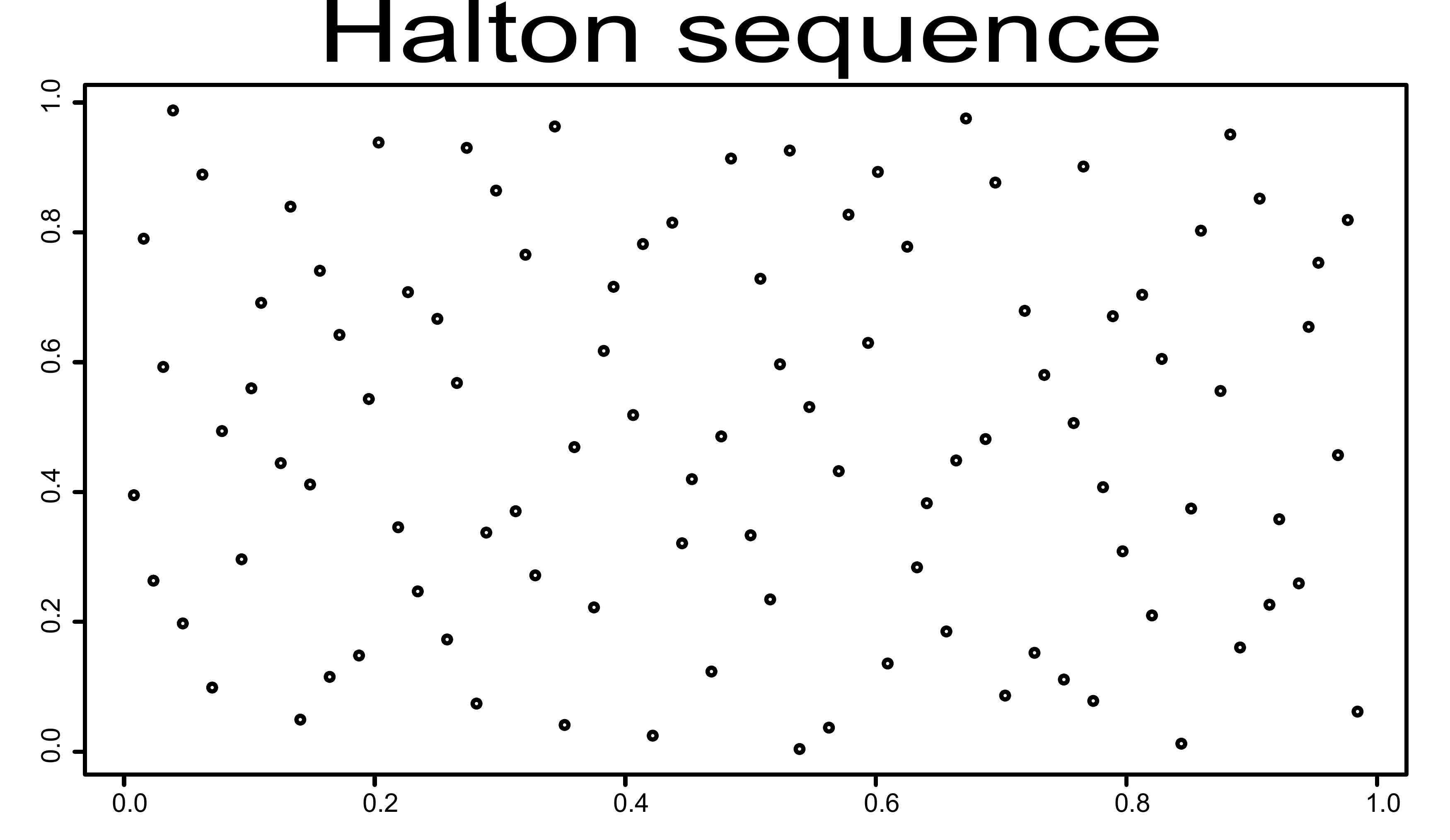}
\includegraphics[scale=0.11]{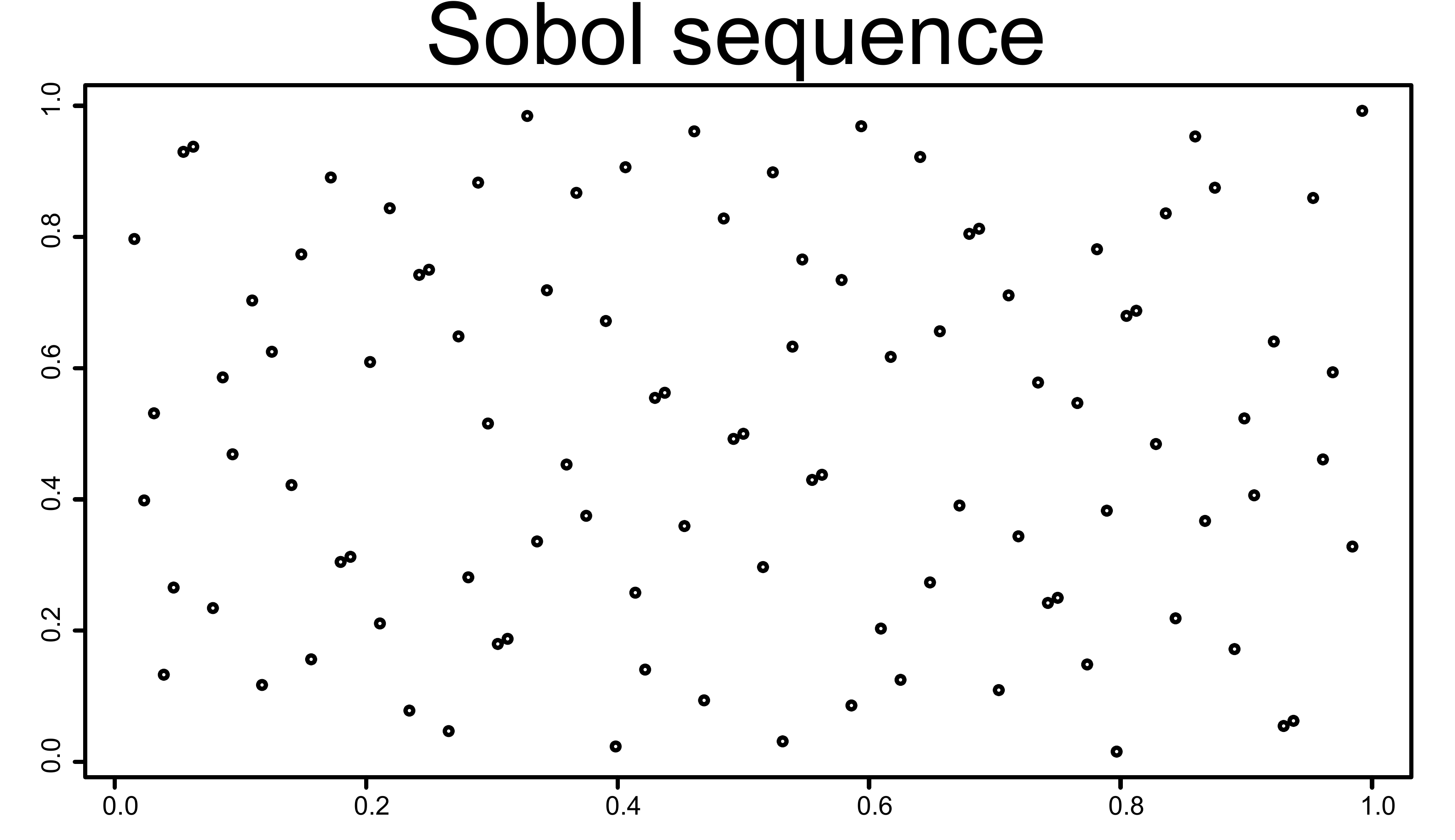}
\includegraphics[scale=0.11]{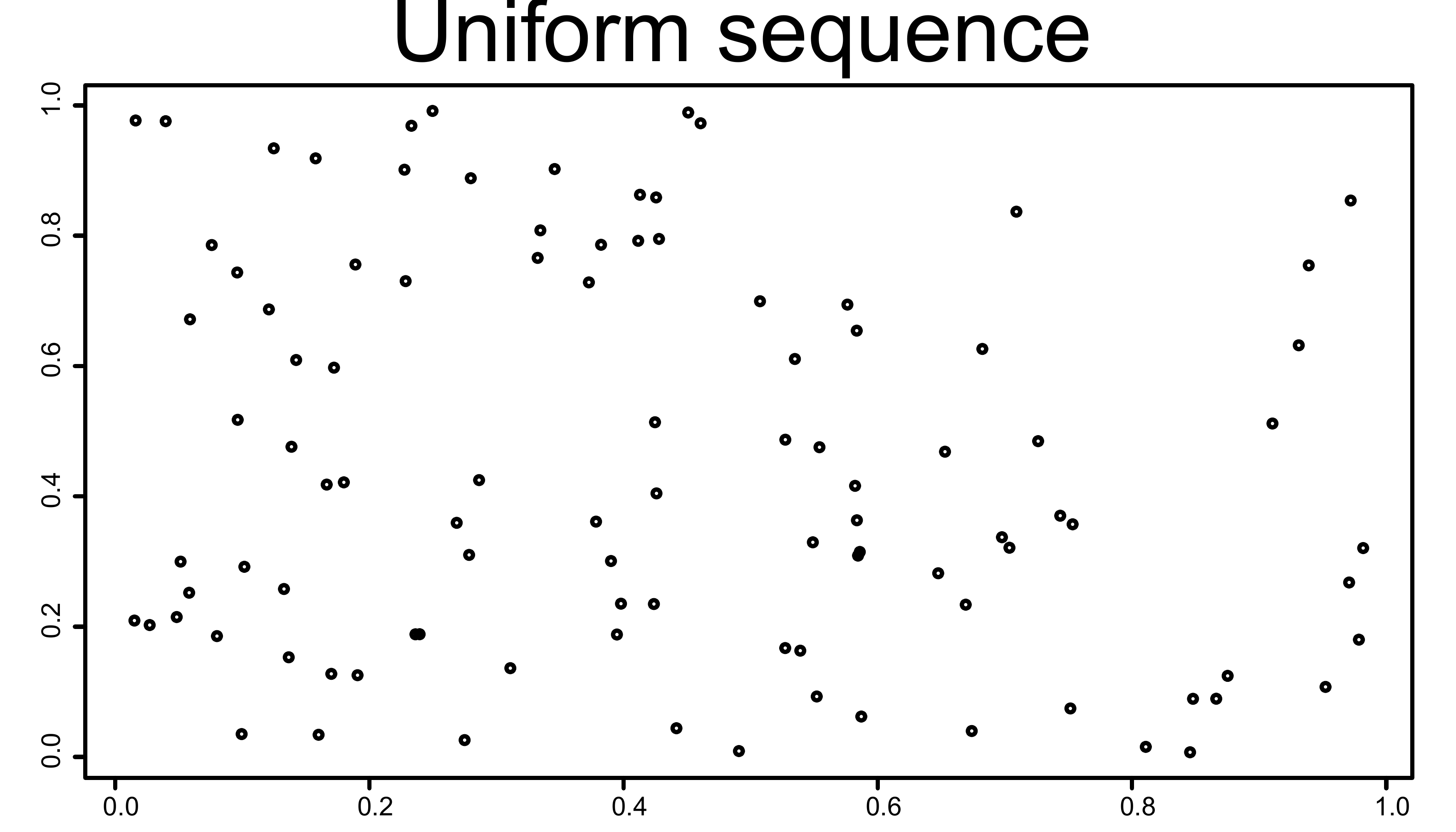}
\caption{Different sequences of points with $\lambda$: 0.2908866, 0.5243953, and 0.6546629 for Halton, Sobol, and uniform sequence respectively}
\label{unif}
\end{figure}

Figure \ref{unif} shows the capability of the coverage measure to compute and quantify the filling of space. Therefore, the use of such measure helps to find the best experimental design regarding the distribution of points.

From this point of view starts the adaptation idea of this measure for the unsupervised feature selection problems. Furthermore, the simplicity of this measure offers a good implementation in a filter algorithm for selecting variables. 

It is important to note that the results giving by this measure are acceptable regarding the selection of the informative feature subset. In addition, it can make use of a parallel CPU computing and a GPU computing to speed up the search procedure.

\section{Unsupervised feature selection based on coverage}
\label{uns}

Numerous techniques exist for implementing redundancy reduction measures. The SFS and SBS are the two commonly used techniques for this purpose. They give acceptable results comparing to the exhaustive search in a short time. The proposed measure can be implemented in any search technique. 
In the remainder of this paper the used search technique is the SFS. The implementation of the proposed measure is described in the following proposition.

\begin{prop}
For all subsets of features, the coverage measure is computed (as it is defined in equation 1). The best subset has the smallest value, regardless to the used search technique.

Since the present work is proposed with a SFS (see algorithm \ref{covr}), the features are added step by step regarding the obtained coverage value.

\end{prop}

\begin{figure}
\centering
\includegraphics[scale=0.13]{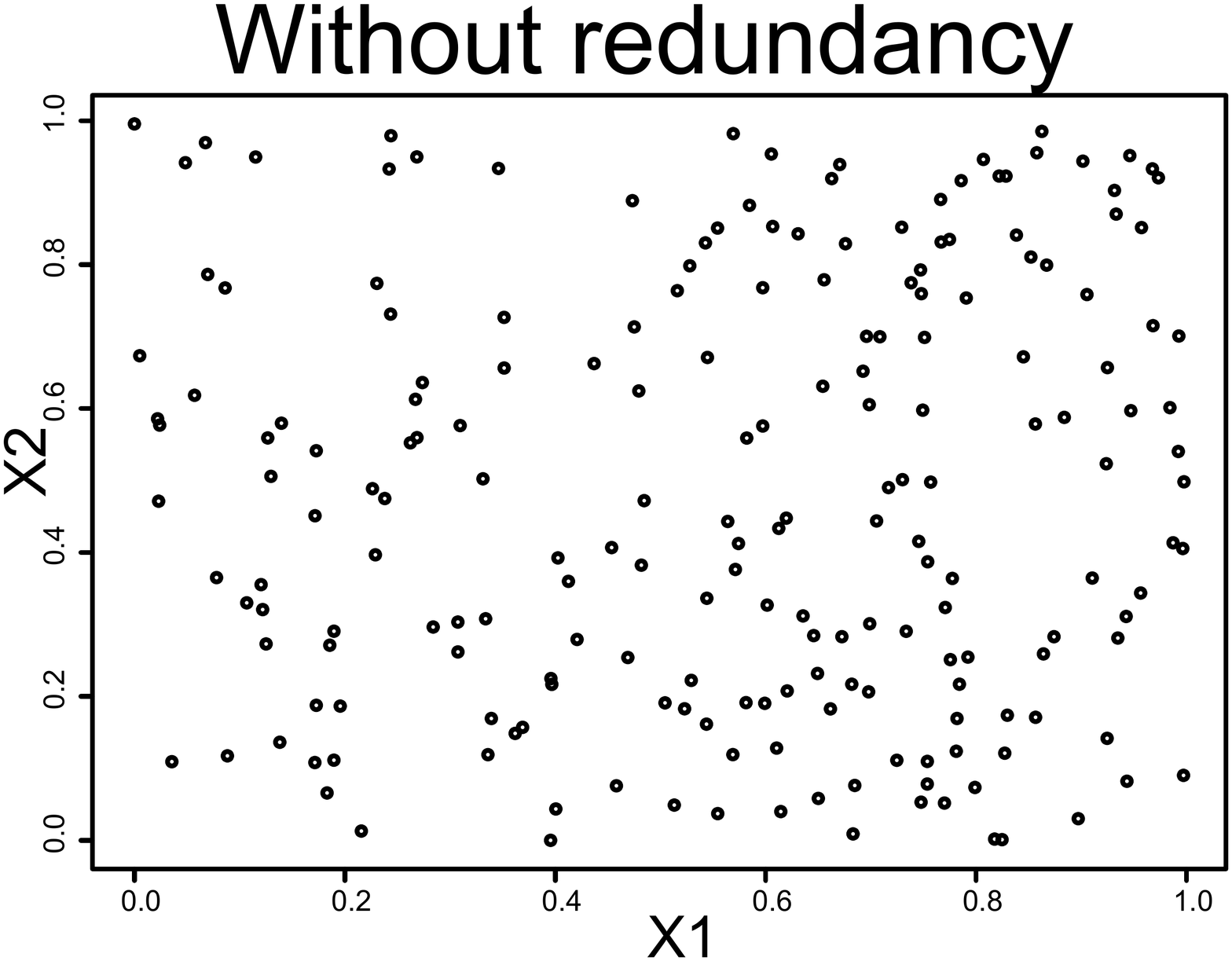}
\includegraphics[scale=0.13]{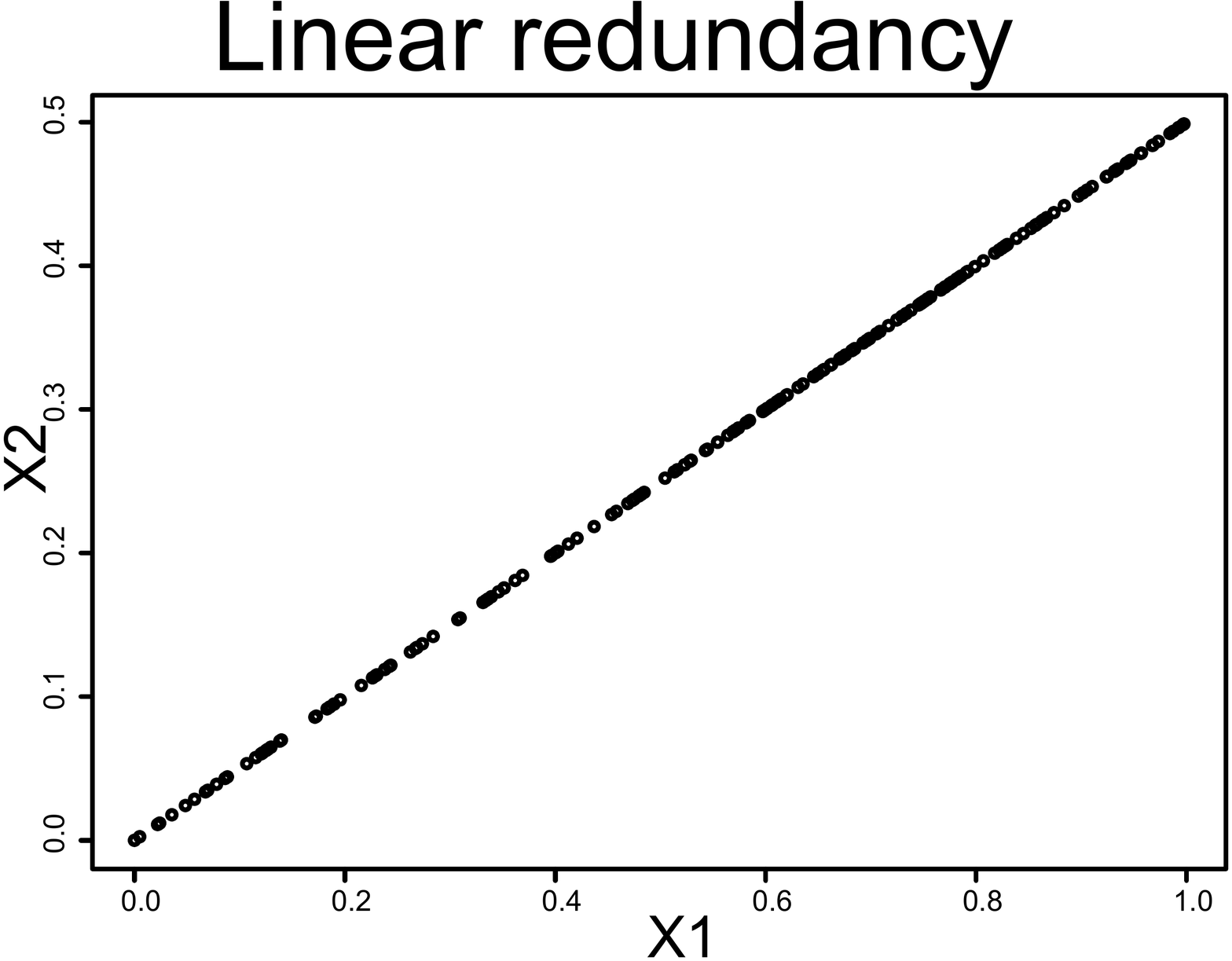}
\includegraphics[scale=0.13]{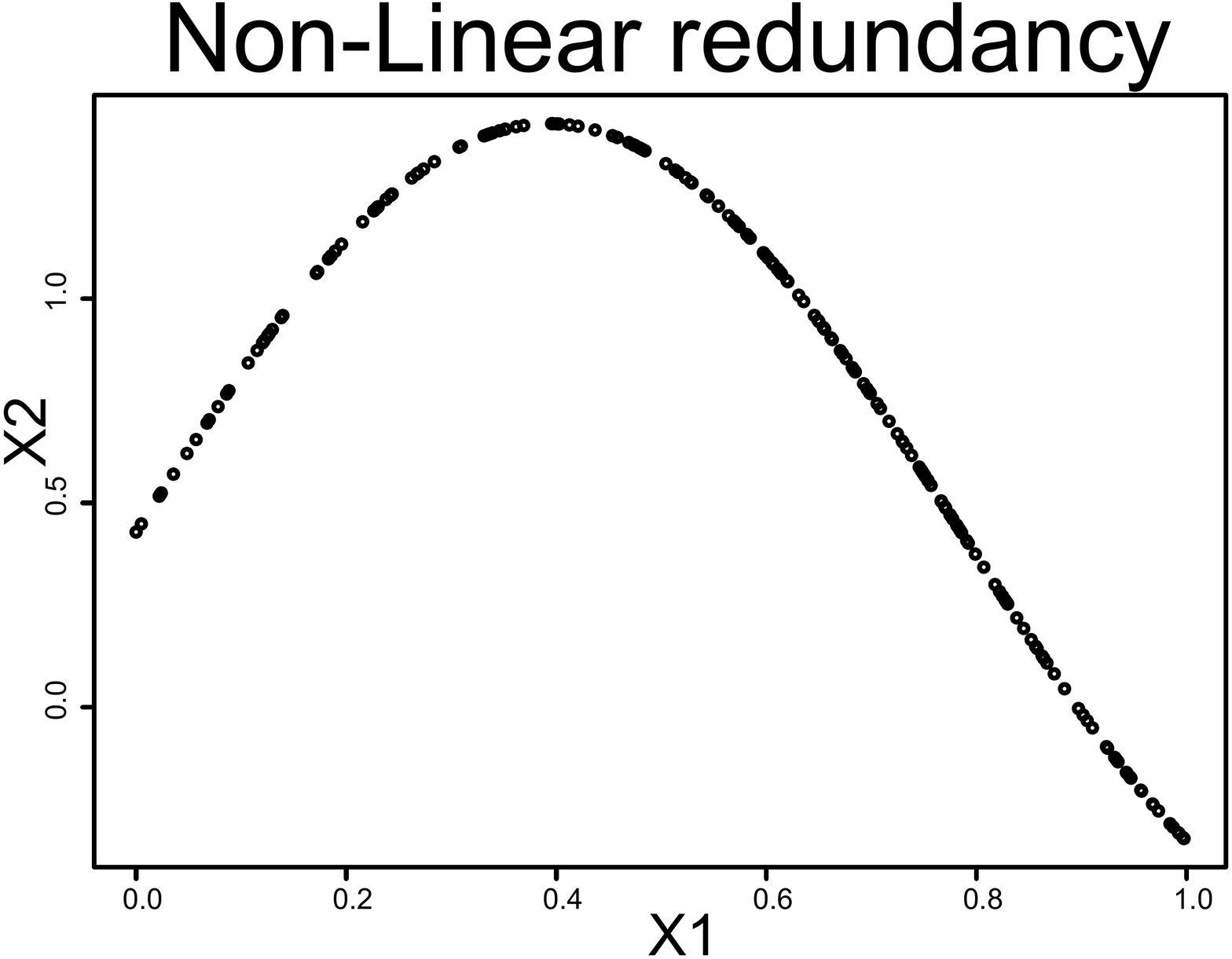}
\caption{Features with $\lambda$: 0.5110806, 1.061314, and 1.106582 for random (or non-redundancy), linear redundancy, and non-linear redundancy respectively}
\label{redun}
\end{figure}

Figure  \ref{redun}  shows clearly that the redundancy is easily detected by this measure, whether it is linear or non-linear redundancy. Besides, the UFS using the (\textsf{UfsCov}) algorithm takes into account the multivariate interactions between selected features. In addition, the \textsf{UfsCov} algorithm does not need extra parameters and does not need a fixed threshold. Therefore, the best subset is the one that gives the smallest coverage measure. Finally, The \textsf{UfsCov} algorithm can be programmed easily in R and MATLAB software.

\begin{algorithm}[t]
\caption{\textsf{UfsCov} algorithm}\label{covr}
\textbf{Input:}  

$\qquad$ Dataset $D$ with $d$ features $X_{1,\ldots, d}$.

$\qquad$ Empty vectors $IdR$ and $CovD$

\textbf{Output:}

$\qquad$ $IdR$ and $CovD$ respectively, the features id and the coverage values.

\begin{algorithmic}[1]
\STATE Rescale data to $[0,1]$.
\FOR{$i = 1 \ \TO \ d$}
\FOR{$j = 1 \ \TO \ (d+1-i)$}
\STATE $\lambda = Coverage(D_{IdR} \: , \: D_j) $
\ENDFOR
\STATE The lowest value of $\lambda$ is stored in $CovD[j]$
\STATE The corresponding id of the lowest $\lambda$ is stored in $IdR[j]$
\ENDFOR\\
\end{algorithmic}
\end{algorithm} 

\section{Experimental case studies}
\label{4}
The simulated and the real world datasets presented in this section are commonly used in several papers on machine learning and feature selection. Moreover, several scenarios of noise injection and shuffling data are proposed to evaluate and to explore the limitation of the \textsf{UfsCov} algorithm.

Further, this section discusses the quality of the obtained results. Finally, the results are verified and evaluated by using random forest algorithm.

\subsection{Simulated case study}
\label{4.1}
The simulated \textit{Butterfly} dataset, introduced in \citep{cran1}, is composed of $8$  features $\{X_1,  X_2,  J_3, J_4, J_5, I_6, I_7, I_8 \}$, where $3$ $\{X_1, X_2,  I_6 \}$ are relevant and contain all the information of the dataset. The remaining $5$ features are constructed basically from $\{X_1, X_2,  I_6 \}$ with linear and non-linear relations. In fact, these $5$ features are  redundant and do not bring new information. (See J. Golay et al. \citep{jango1}). 

\begin{figure}
\includegraphics[scale=0.4]{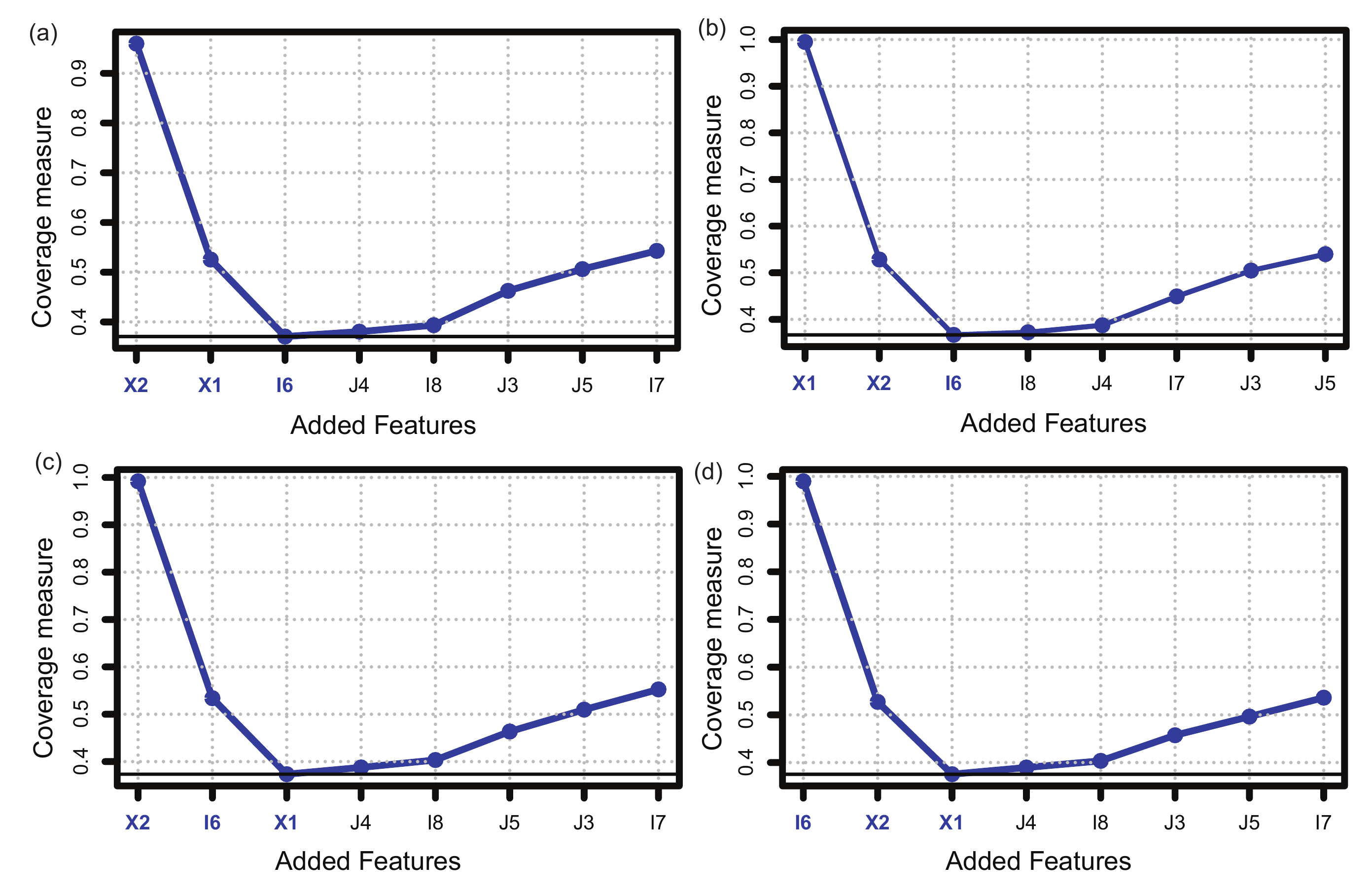}
\caption{The butterfly dataset results, with: (a) $N=1000$, (b) $N=2000$, (c) $N=5000$, (d) $N=10000$. The algorithm chooses the correct subset of feature regardless of the number of points (the minimum is reached at 3 features).}
\label{but}
\end{figure}

Figures \ref{but} show the results for the Butterfly datasets with different number of $N$ points. The results show that the \textsf{UfsCov} algorithm finds easily the three important features $\{X_1, X_2, I_6 \}$ regardless of the number of points used to generate the Butterfly dataset. The minimum value of the coverage measure is reached at the correct subset.

\subsection{Noise injection}
The robustness of \textsf{UfsCov} is evaluated against noise. In fact, several experiments of noise injection were performed for two different scenarios. The first one consists in injecting noise to all features of the \textit{Butterfly} dataset. The second one consists to corrupt only the redundant features ($J_3, J_4, J_5 , I_7, I_8 $). A Gaussian noise is used with a mean fixed at $0$ and a standard deviation set at: $1\%, 5\%, 10\%, 20\%$, and $50\%$ of the original standard deviation of feature.

The objective of these experiments is to see if \textsf{UfsCov} can detect an existing redundancy in data corrupted by a Gaussian noise. Furthermore, it is important to find out the limitation of this algorithm against noise and at what level.

\begin{figure}
\includegraphics[width=\linewidth]{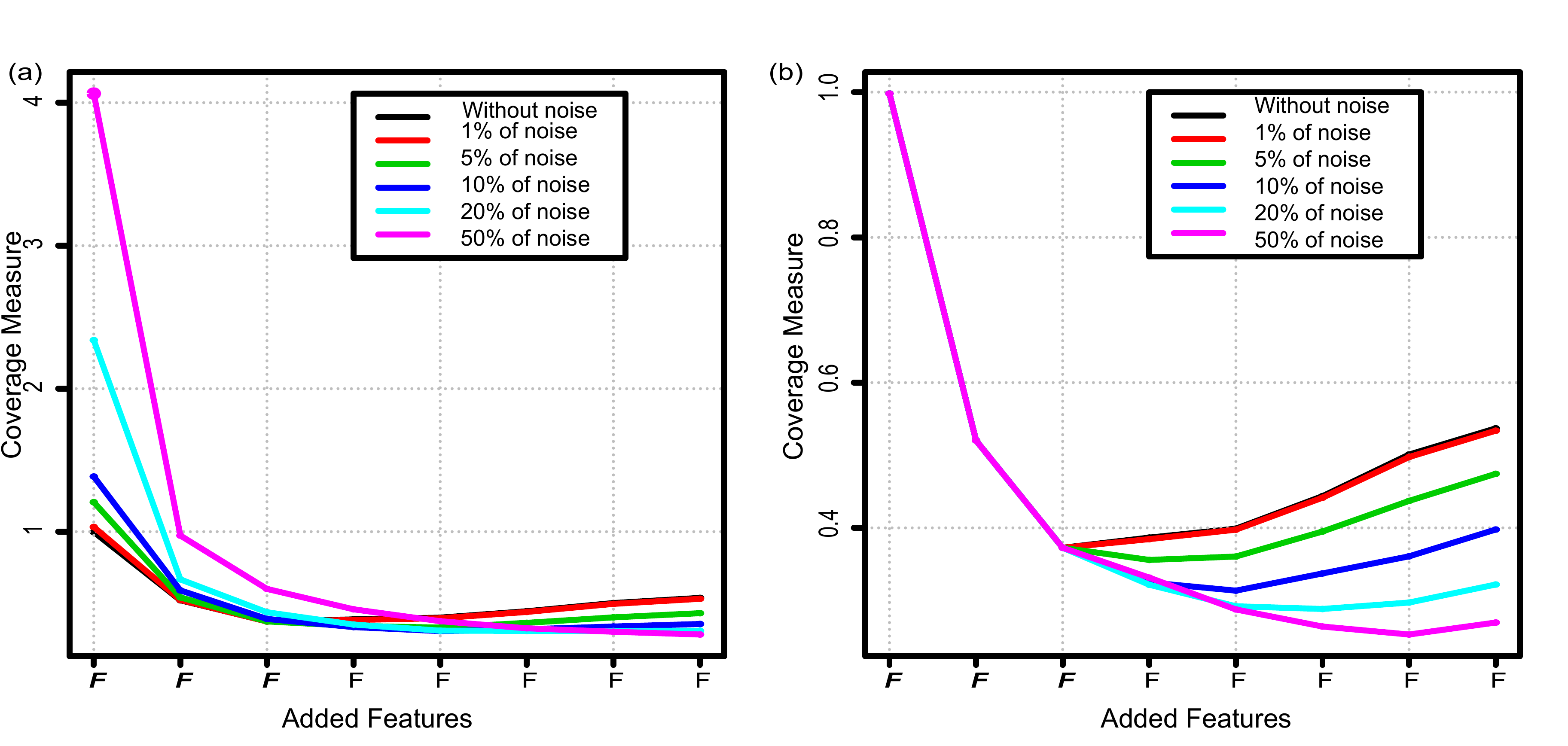}
\caption{Noise injection: (a) all features are corrupted with a Gaussian noise, (b) only redundant feature are injected with a Gaussian noise.}
\label{noise}
\end{figure}

Figure \ref{noise} shows the two proposed scenarios of noise injection. Figure \ref{noise}.b presents the reaction of \textsf{UfsCov} with corrupted redundant features, at different levels of noise. The algorithm is still robust and detects the important features. However, at $50\%$ of noise, the minimum value of the coverage is not indicating the correct subset of features, which is normal for such level of noise. On the other hand, the algorithm gives at least a correct ranking of features regarding the importance and the provided information of each feature (see table \ref{somenoise}). Therefore, it can be concluded that the \textsf{UfsCov} algorithm is robust against noise.

\begin{table}
\centering 
\begin{tabular}{|c|c|c|c|c|c|}
\hline 
Without noise  & $1\%$    &  $5\%$   & $10\%$  & $20\%$  & $50\%$ \\ 
\hline 
\hline 
\textbf{$X_1$}   & \textbf{$X_1$}     &  \textbf{$X_2$}    & \textbf{$X_1$}  & \textbf{$X_1$}  & \textbf{$I_6$} \\ 

\textbf{$I_6$}  & \textbf{$I_6$}   &  \textbf{$I_6$}    & \textbf{$I_6$}  & \textbf{$X_2$} & \textbf{$X_2$} \\ 

\textbf{$X_2$}  & \textbf{$X_2$}   &  \textbf{$X_1$}     & \textbf{$X_2$} & \textbf{$I_6$} & \textbf{$X_1$}  \\ 

$I_8$ & $I_8$    &  $I_8$    & $I_8$  & $I_8$ & $J_4$ \\ 

$J_4$  & $J_4$    &  $J_4$    & $J_4$  & $J_4$ & $J_5$ \\ 

$I_7$  & $I_7$    &  $I_7$     & $I_7$   & $I_7$  & $J_3$ \\ 

$J_5$  & $J_5$    & $J_5$  & $J_5$  & $J_5$  & $I_8$ \\ 

$J_3$  & $J_3$    &  $J_3$    & $J_3$  & $J_3$ & $I_7$ \\ 
\hline 
\end{tabular}
\caption{The added features in the SFS technique (here, the Gaussian noise is injected to all features).}
\label{somenoise}
\end{table}

\begin{table}
\centering 
\begin{tabular}{|c|c|c|c|c|c|}
\hline 
Without noise  & $1\%$    &  $5\%$    & $10\%$   & $20\%$  & $50\%$  \\ 
\hline 
\hline 
\textbf{$X_1$}  & \textbf{$X_1$ }   & \textbf{$X_1$}    & \textbf{$X_1$}  & \textbf{$X_1$ }& \textbf{$X_1$} \\ 

\textbf{$I_6$}  & \textbf{$I_6$}    &  \textbf{$I_6$}    & \textbf{$I_6$}  & \textbf{$I_6$} & \textbf{$I_6$} \\ 

\textbf{$X_2$}  & \textbf{$X_2$}    &  \textbf{$X_2$}     & \textbf{$X_2$}   & \textbf{$X_2$}  & \textbf{$X_2$}  \\ 

$I_8$  & $I_8$    &  $I_8$    & $I_8$  & $I_8$ & $I_8$ \\ 

$J_4$ & $J_4$    & $J_4$     & $J_4$  & $J_4$  & $J_4$  \\ 

$I_7$ & $I_7$    &  $I_7$    & $J_5$  & $J_5$ & $J_5$ \\ 

$J_5$  & $J_5$    &  $J_5$    & $I_7$  & $I_7$ & $I_7$ \\ 

$J_3$  & $J_3$    &  $J_3$    & $J_3$ & $J_3$ & $J_3$ \\ 
\hline 
\end{tabular}
\caption{The added features in the SFS technique (here, the Gaussian noise is injected to all features).}
\label{allnoise}
\end{table}

\subsection{Shuffling features}

\begin{figure}
\includegraphics[width=\linewidth]{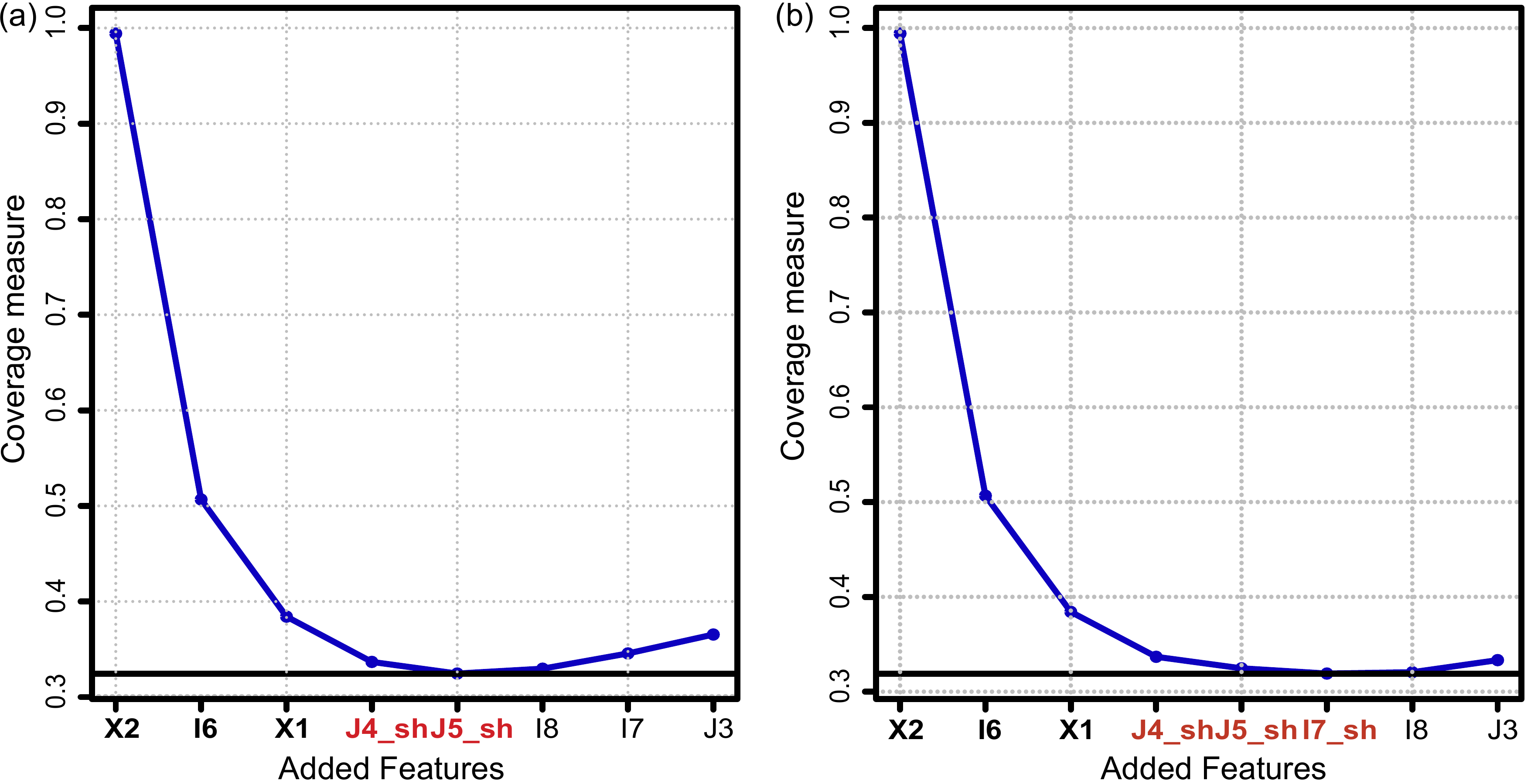}
\caption{Shuffling features: (a) $J_4$ and $J_5$ are shuffled, (b) the features: $J_4$, $J_5$, and $I_7$ are shuffled .}
\label{shuffling}
\end{figure}

In addition to injecting noise in data, shuffling of features can be an interesting experiment to evaluate the \textsf{UfsCov} algorithm. This operation was carried out with two scenarios: at the beginning, only two redundant features are shuffled ( $J_4$, $J_5$). Then, three redundant features are shuffled. The results were expected, since the shuffling destroys the linear or non-linear relation between features. In fact, this can reduce the redundancy. As figure \ref{shuffling} presents, \textsf{UfsCov} selected features with relevant information (which are not redundant).
 
\subsection{Benchmark case studies}

Benchmark case studies \cite{UCI, data0} are also used to test the \textsf{UfsCov} algorithm. The datasets used in this work are: Parkinson, PageBlocks, Ionosphere,  and COIL20 \cite{data0}. Table \ref{data} describes these datasets and the number of selected features for each dataset. 

\begin{table}
\centering
\begin{tabular}{|lccr|}
\hline 
Data  & Number of instances    &  Number of features & Selected features \\ 
\hline 
\hline 
Parkinson &   $195$  & $22$ & $7$   \\ \hline 

PageBlocks   & $5393$ & $10$ & $3$ \\ \hline 

Ionosphere  & $350$  & $34$  & $10$ \\  \hline 

COIL20     & $1440$  & $1024$ & $98$ \\ \hline 

\end{tabular}
\caption{Description of the used datasets and summary of the results obtained by the \textsf{UfsCov} algorithm.}
\label{data}
\end{table}

\begin{figure}
\centering
\includegraphics[width=\linewidth]{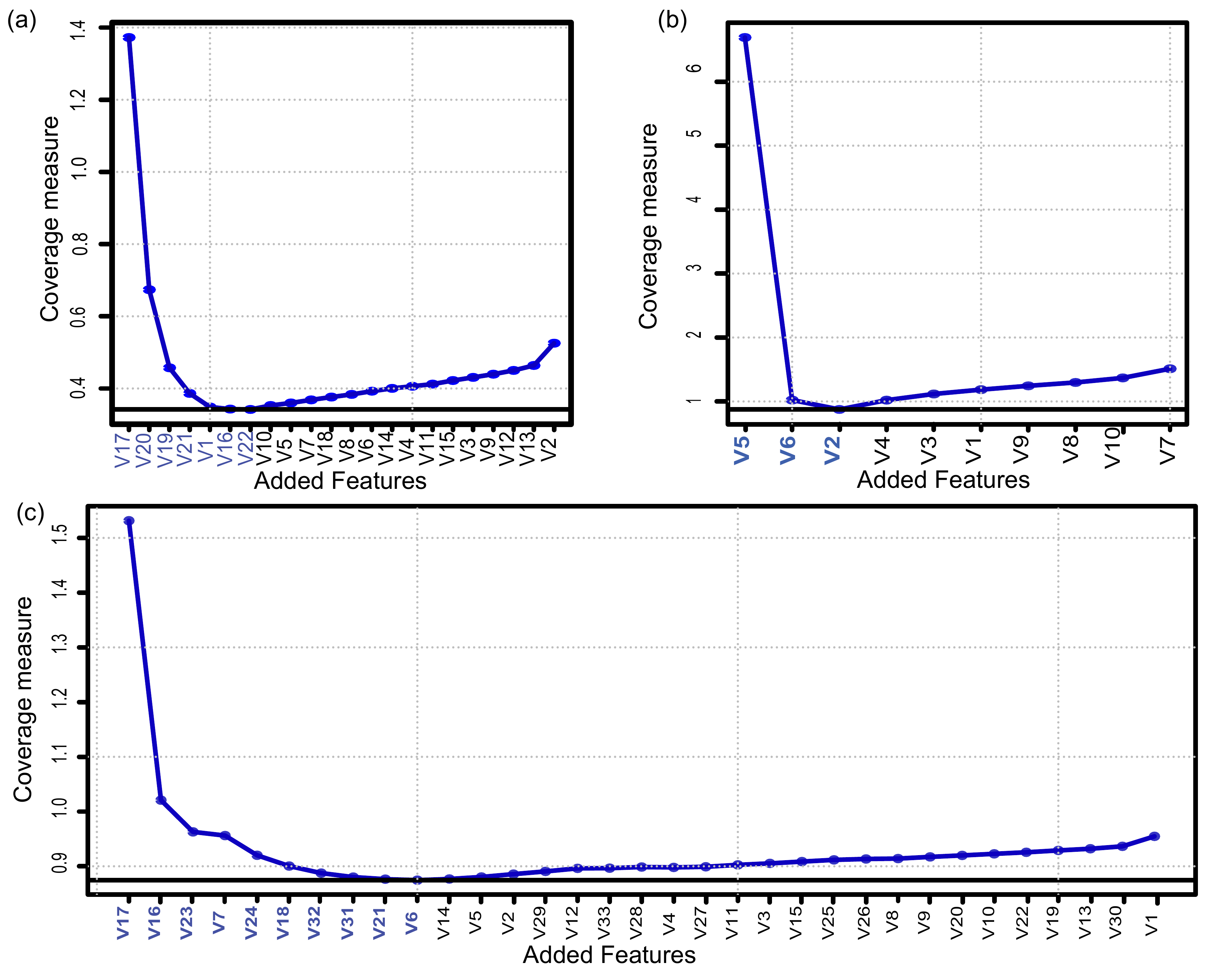}
\caption{Results of the \textsf{UfsCov} algorithm for: (A) Parkinson, (B) PageBlocks, (C) Ionosphere. The selected subset of features provides the minimum value of coverage measure}
\end{figure}

\begin{figure}
\centering
\includegraphics[width=\linewidth]{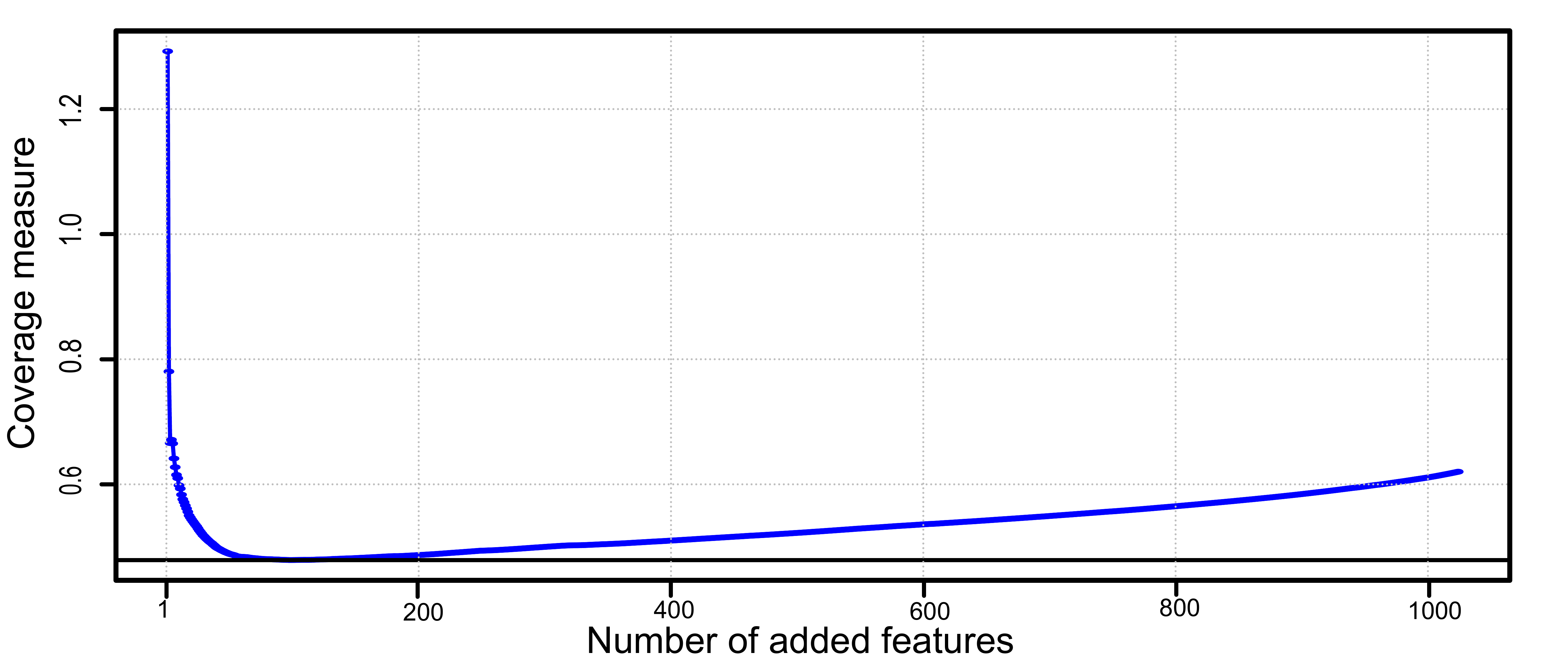}
\caption{Results of the \textsf{UfsCov} algorithm for the Coil20 dataset. The minimum is reached with $98$ features.}
\end{figure}

\subsubsection{Results and discussions}
In addition to applying the \textsf{UfsCov} algorithm on simulated and real world datasets, this subsection discusses the evaluation of the results. Here, random forest algorithm is used as a classifier for the four datasets used above (Parkinson, PageBlocks, Ionosphere,  and COIL20).

The used procedure of testing with random forest is applied once with all features of the datasets and once with only selected features. The procedure can be summarised as follows:

\begin{itemize}
\item the data were split into training and testing sets ($80\%$ for training and $20\%$ for testing);
\item the training set was used to find the optimal parameters of random forest (the number of trees and the number of predictors). Furthermore, the training step was performed by using a $10$-fold cross-validation;
\item a random forest model was generated with the optimal parameters found above (previous step), and then applied to classify the testing set. Two classification evaluation metrics are used:

\begin{itemize}
\item \textbf{the overall accuracy} of classification is computed with the following formula:
\begin{equation}
OA= \frac{1}{n} \sum^{n}_{i=1}I(y_{i} = \hat{y}_{i})
\end{equation}

where $\hat{y}_{i}$ is the predicted class label for the $i$th observation using the random forest model. And $I(y_{i} = \hat{y}_{i})$ is an indicator variable with:
\begin{equation}
I(y_{i} = \hat{y}_{i})=\left\{\begin{array}{rcl}
1 \qquad \qquad  \; \; \mbox{if correctly classified} \\
\\
0  \qquad  \; \;  \mbox{otherwise (misclassified)}
\end{array}\right.  
\end{equation}
Therefore, the OA formula computes the fraction of correct classifications, which means that the best classification has the highest overall accuracy.

\item \textbf{Cohen's Kappa coefficient} \textsf{k}  \cite{kappa1} is also used to compare the classification results of random forest. The Kappa evaluation metric is computed on the test subset by using the following formula:

\begin{equation}
k=\frac{n \sum_{c}T_{c}- \sum_{c}P_{c}}{n^{2}- \sum_{c} G_{c} P_{c}}
\end{equation}
where $T_{c}$ indicates the number of correctly classified samples for class $c$; and $n$ is the number of data points in the test subset. $G_{c}$ and $P_{c}$ are the size of samples for the class $c$ and the samples classified in the same class $c$.
\end{itemize}

\end{itemize}

During the evaluation, random forest algorithm was repeated $20$ times. Tables \ref{rdata} and \ref{kdata} illustrate the obtained results with the overall accuracy and the Kappa coefficient respectively.
\begin{table}
\centering 
\begin{tabular}{|lcr|}
\hline 
Data  & All features    &  Selected features \\ 
\hline 
\hline 

Parkinson &   $0.916 \: (0.042)$  & $ 0.918 \: (0.039) $  \\\hline 

PageBlocks   & $0.977 \: (0.004)$ &  $ 0.951 \: (0.006)$\\\hline 

Ionosphere  & $0.913 \: (0.024)$  &  $ 0.915 \: (0.032) $ \\  
\hline 
COIL20     & $0.995 \: (0.001)$  & $0.996 \: (0.002)$ \\
\hline 
\end{tabular}
\caption{In percent: random forest classification errors (20 repetitions with random splits) and the standard deviation as well.}
\label{rdata}
\end{table}

\begin{table}
\centering 
\begin{tabular}{|lcr|}
\hline 
Data  & All features    &  Selected features \\ 
\hline 
\hline 

Parkinson &   $0.7068  \: (0.1140)$  & $ 0.7374 \: (0.109) $  \\\hline 

PageBlocks   & $0.8518 \: (0.0221)$ &  $ 0.7168 \: (0.0340)$\\\hline 

Ionosphere  & $0.8289 \: (0.0623)$  &  $ 0.8242 \: (0.0722) $ \\  
\hline 
COIL20     & $0.9964 \: (0.0004)$  & $0.9945 \: (0.0012)$ \\
\hline 
\end{tabular}
\caption{Mean Kappa coefficient (over 20 repetitions with random splits) and the standard deviation as well.}
\label{kdata}

\end{table} 
The obtained results presented in tables \ref{rdata} and \ref{kdata} show that the \textsf{UfsCov} algorithm kept only the relevant informative features. In fact, it reduces the existing redundancy in data. Therefore, the proposed filter algorithm could be an interesting tool to minimise redundancy in data.

\subsection{Environmental case studies}
This section shows the potential of the proposed unsupervised feature selection algorithm on environmental data. In fact, the algorithm is applied on Permafrost data and the Indian Pines hyperspectral image.

\subsubsection{Permafrost case study}
The data were collected in the Alp Mountains of Switzerland. $26$ features (excluding the XY coordinates) are used to predict Permafrost presence or absence. For more details on the study, including more complete references and more information about the collected features, see N. Deluigi et al. \citep{nicola}.
\begin{figure}
\includegraphics[scale=0.4]{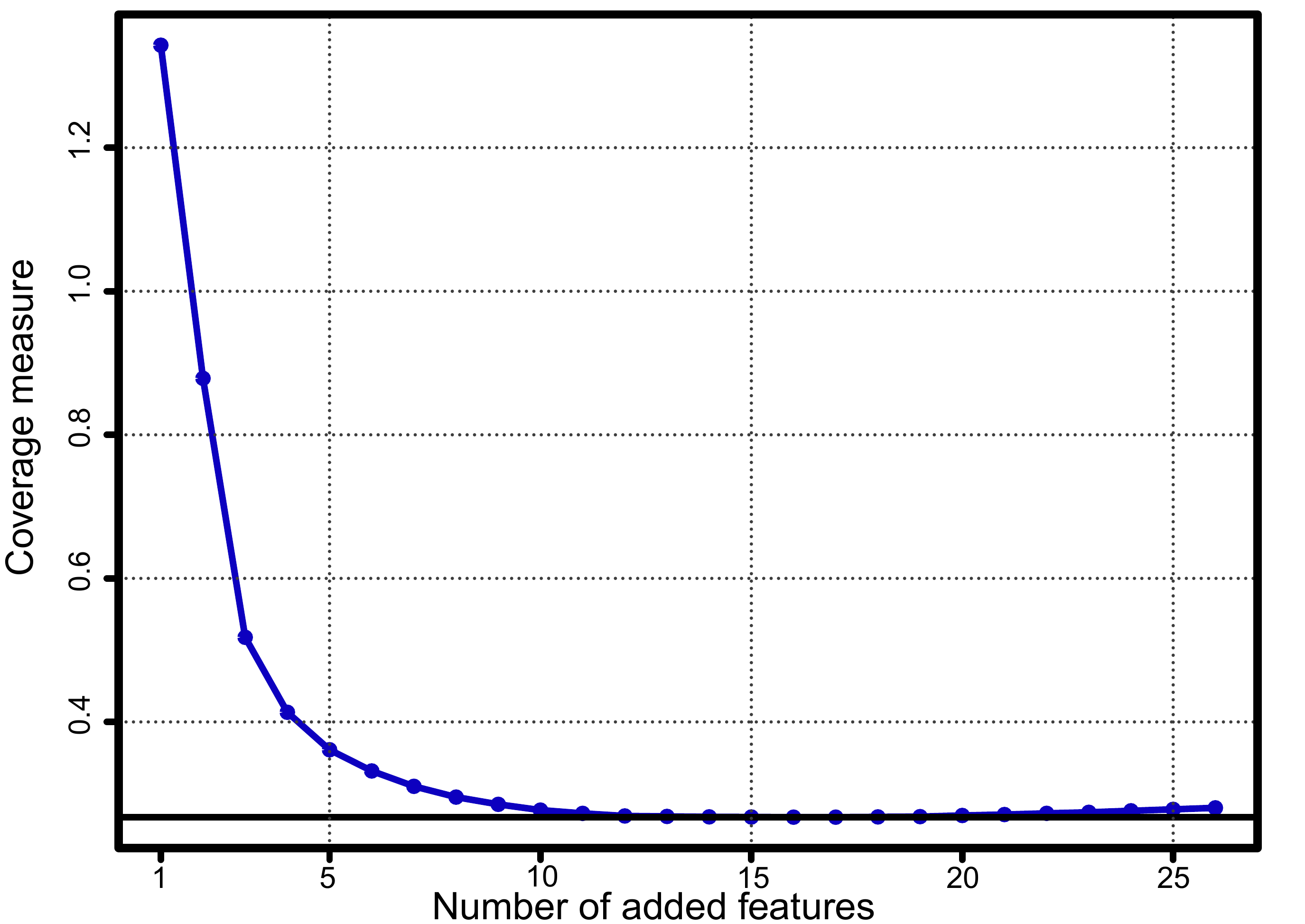}
\caption{Results of the UfsCov algorithm applied on the Permafrost data. The minimum of the coverage value is reached at $17$ features.}
\label{perma1}
\end{figure}

\begin{figure}
\includegraphics[scale=0.21]{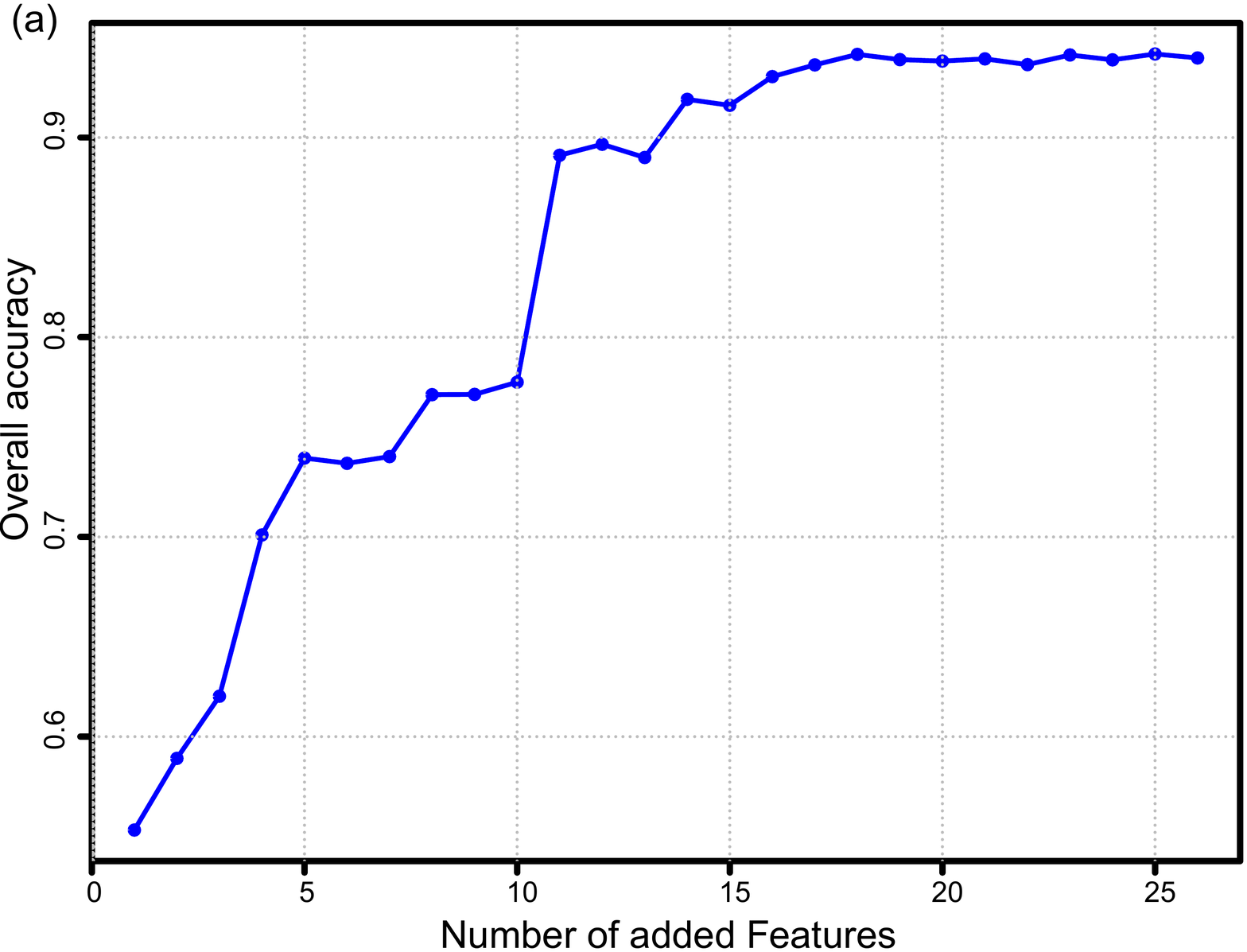}
\includegraphics[scale=0.21]{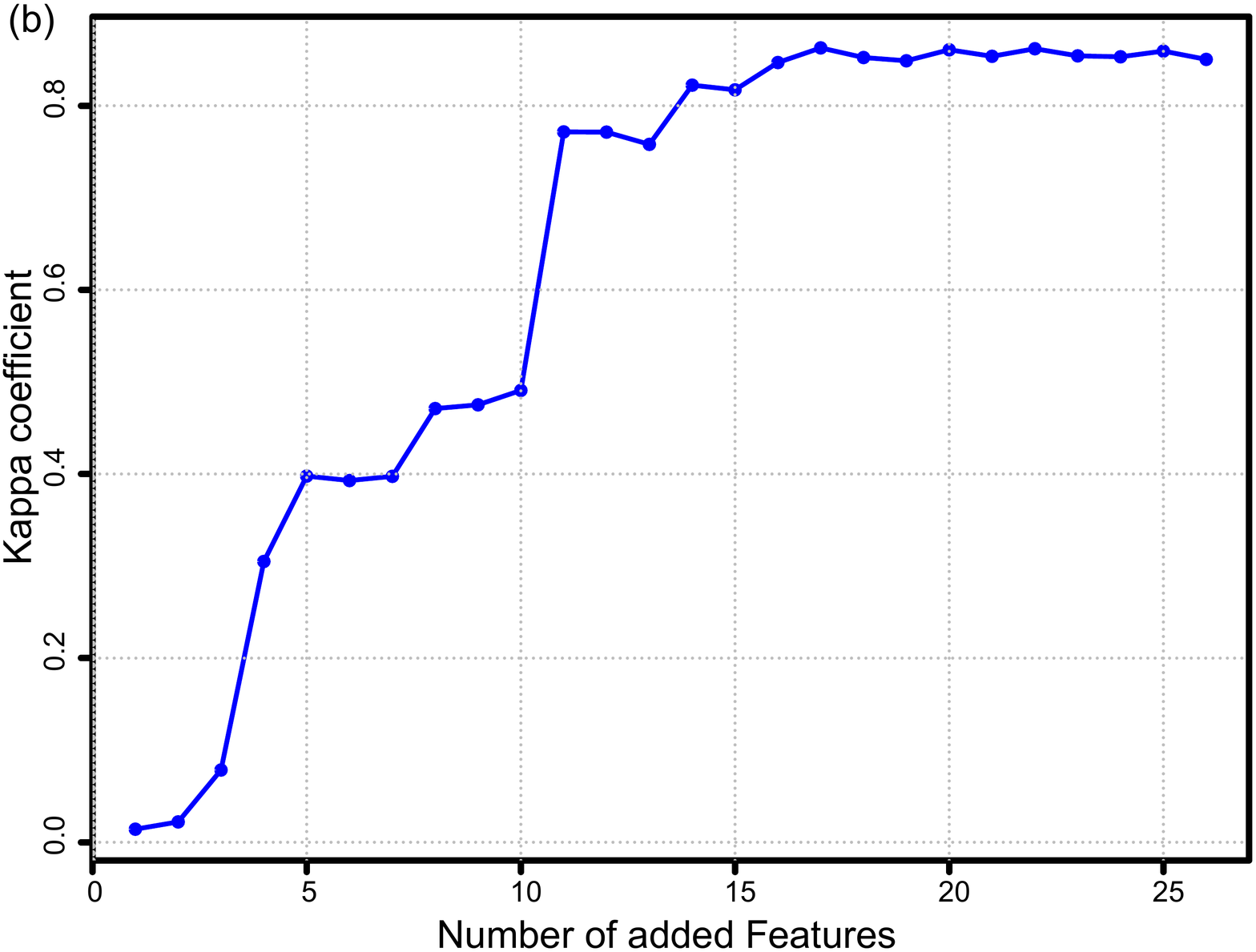}
\caption{Random forest results for each step of the UfsCov (a) the overall accuracy, (b) the Kappa coefficient. (Permafrost dataset)}
\label{perma2}
\end{figure}

Figure \ref{perma1} presents the unsupervised feature selection results. The minimum of the coverage measure is reached at $17$ features. Furthermore, the given result is evaluated by using random forest algorithm. Table \ref{permat} shows the results of random forest, with all features and with only the selected features. The classification accuracy and the Kappa coefficient are shown in figure \ref{perma2}. In this figure, random forest is applied after each step of UfsCov algorithm. 
\begin{table}
\centering 
\begin{tabular}{|l|c|c|}
\hline
Features & Accuracy & Kappa metric \\
\hline
All features ($26$)& $0.9301 \; (0.0002)$ & $0.848 \; (0.0008)$ \\
\hline
Selected features ($17$)& $0.9339 \; (0.002)$ & $0.867 \; (0.0087)$ \\
\hline
\end{tabular}
\caption{Random forest errors and the standard deviation after $20$ repetitions with different splitting (Permafrost dataset).}
\label{permat}
\end{table}

\subsubsection{Indian Pines image}
\begin{figure}
\frame{\includegraphics[scale=0.23]{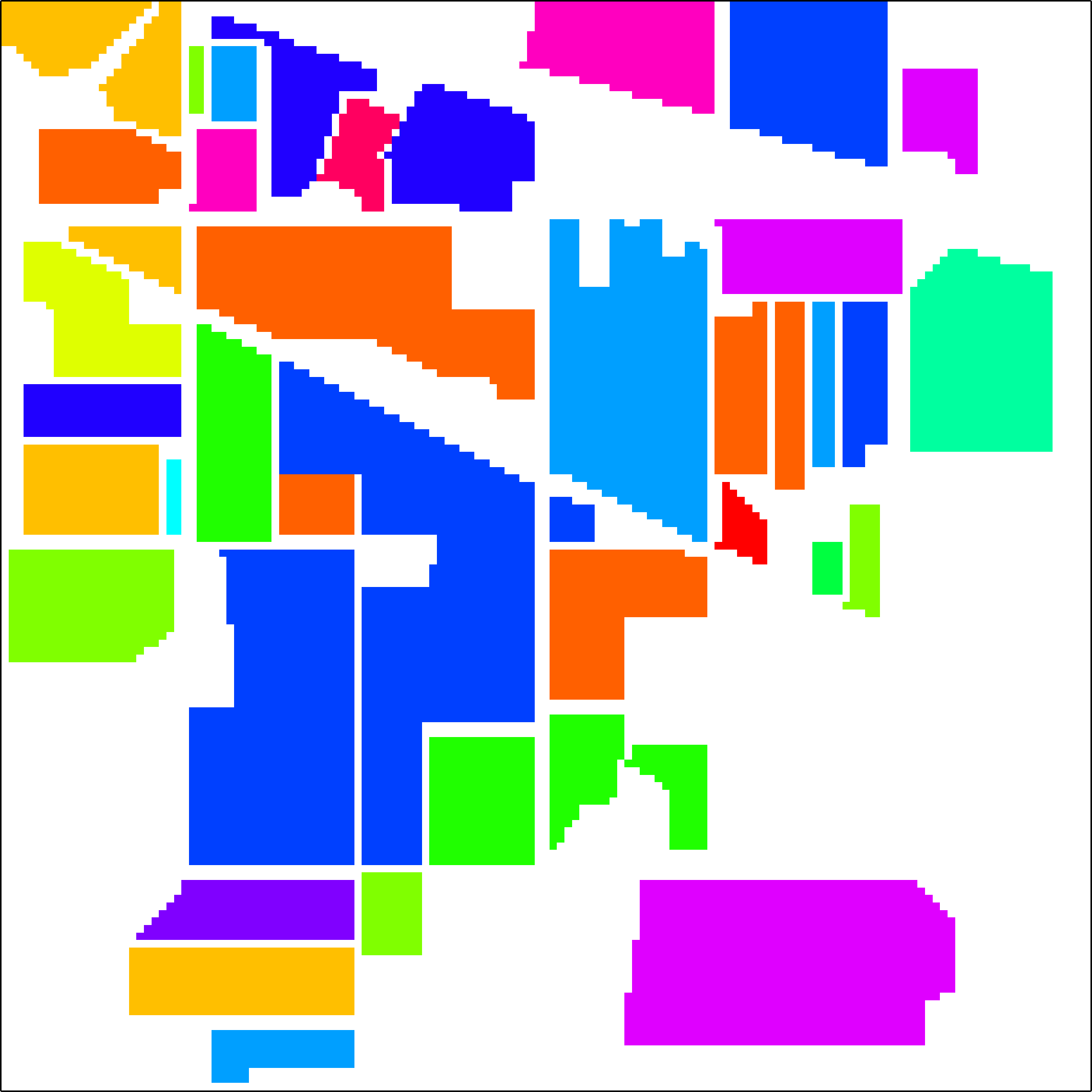}}
\includegraphics[scale=0.23]{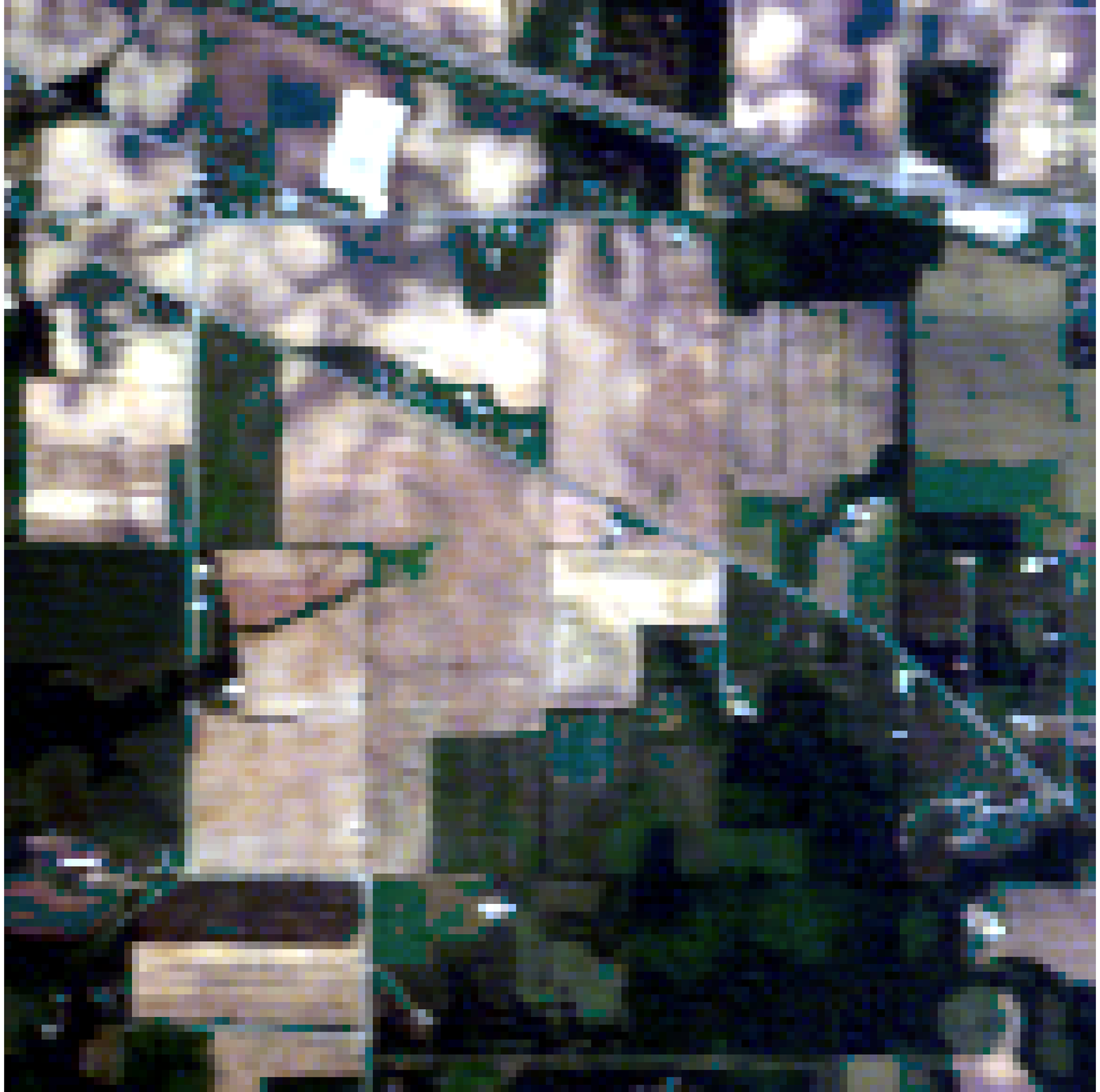}
\caption{The Indian Pine Site image with a size of $145$ x $145$ pixels
and the ground truth of the Indian Pine Site dataset.}
\label{ipines}
\end{figure}

The image was captured by Airborne Visible/Infrared Imaging Spectrometer (AVIRIS) sensor in the Northwest Indiana, on June 12, 1992.  The Indian Pines scene contains agricultural and forested region (figure \ref{ipines} ). The data consist of  $145$ x $145$ pixels and $220$ spectral bands with a spatial resolution of $20$ m/pixel \cite{PURR1947}.

In this work only $200$ bands are used for the experiments, after removing $20$ noisy bands ($104-108$, $150-163$, $220$) due to water absorption. The present case study of hyperspectral image shows that the \textsf{UfsCov} algorithm is able to deal with remote sensing problems. Furthermore, the proposed algorithm help to manage high dimensional datasets (more than $100$ features).

\begin{figure}
\includegraphics[width=\linewidth]{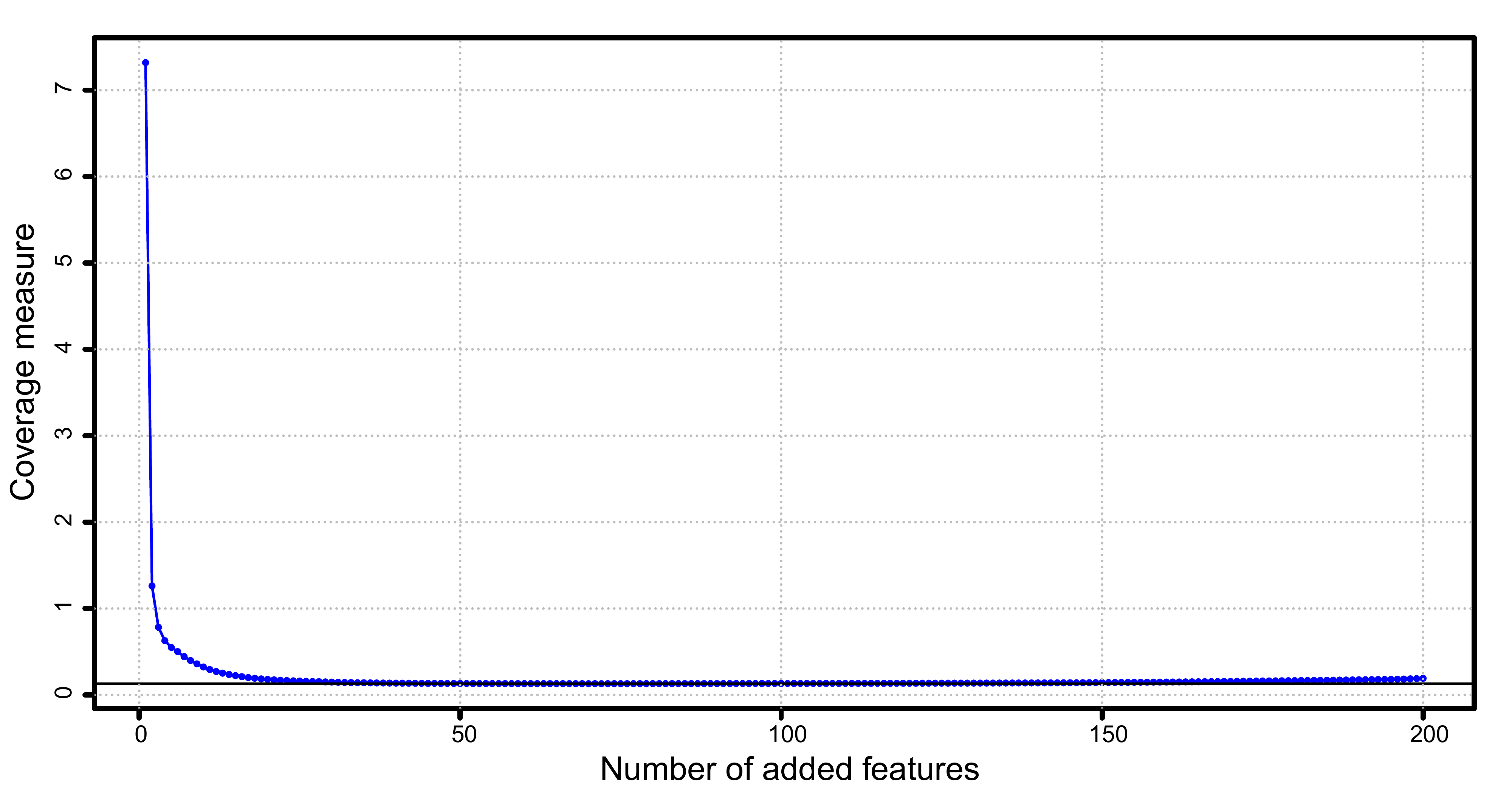}
\caption{Results of the UfsCov algorithm applied on the Indian Pines image. The minimum of the coverage value is reached at $69$ features }
\label{indian}
\end{figure}

Figure \ref{indian} shows the results of the proposed algorithm. The minimum value is reached with the $69$ features. Table \ref{indiana} compares the difference between the two random forest models, with all features and with only the selected features.

\begin{table}
\centering 
\begin{tabular}{|l|c|c|}
\hline
Features & Accuracy & Kappa metric \\
\hline
All features ($200$ bands)& $0.870\; (0.0055)$ & $0.852 \; (0.0066)$ \\
\hline
Selected features ($69$ bands)& $0.844 \; (0.0037)$ & $0.822 \; (0.0036)$ \\
\hline
\end{tabular}
\caption{Random forest errors and the standard deviation after $20$ repetitions with different splitting (for the Indian Pines image).}
\label{indiana}
\end{table}

\begin{figure}
\includegraphics[scale=0.21]{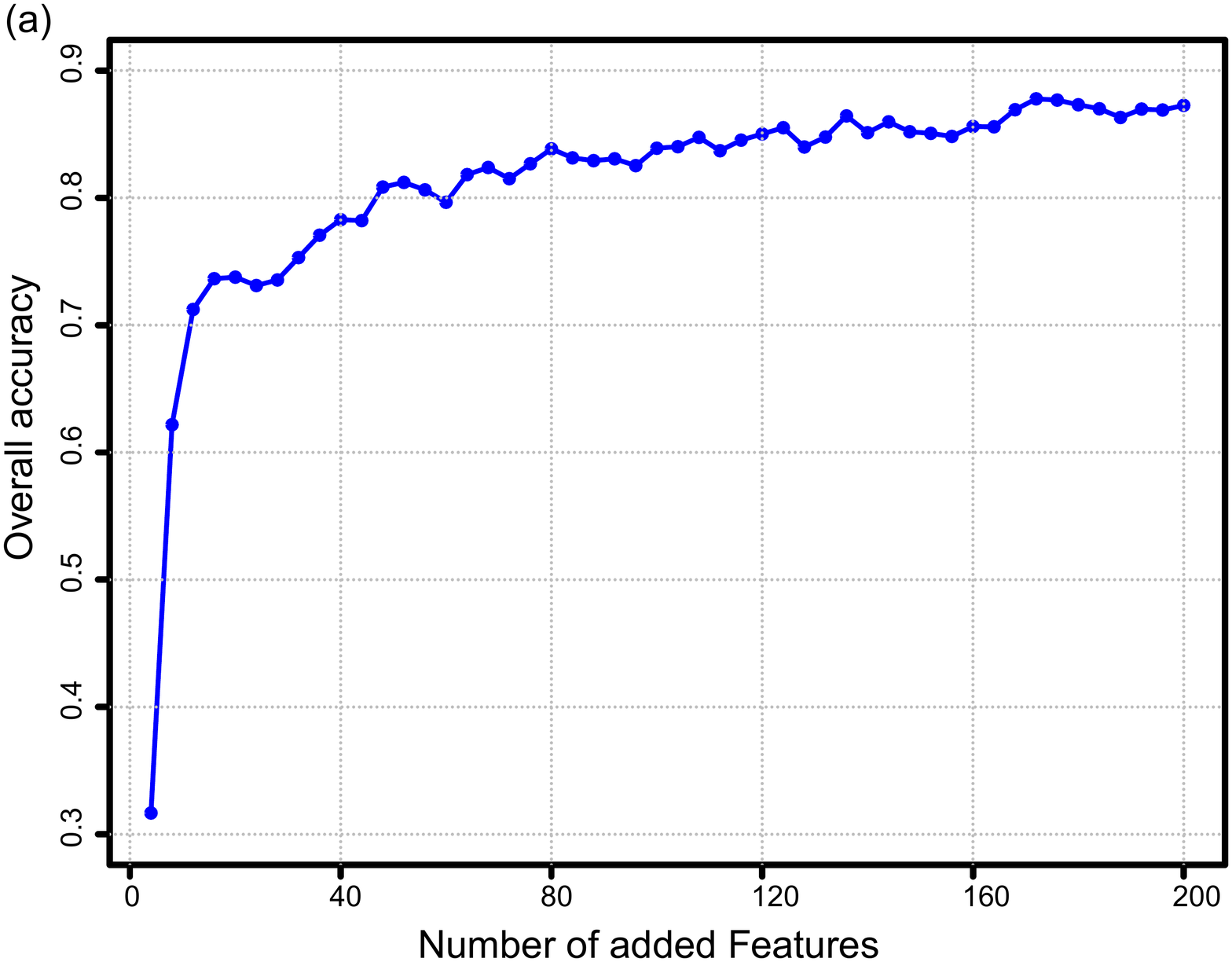}
\includegraphics[scale=0.22]{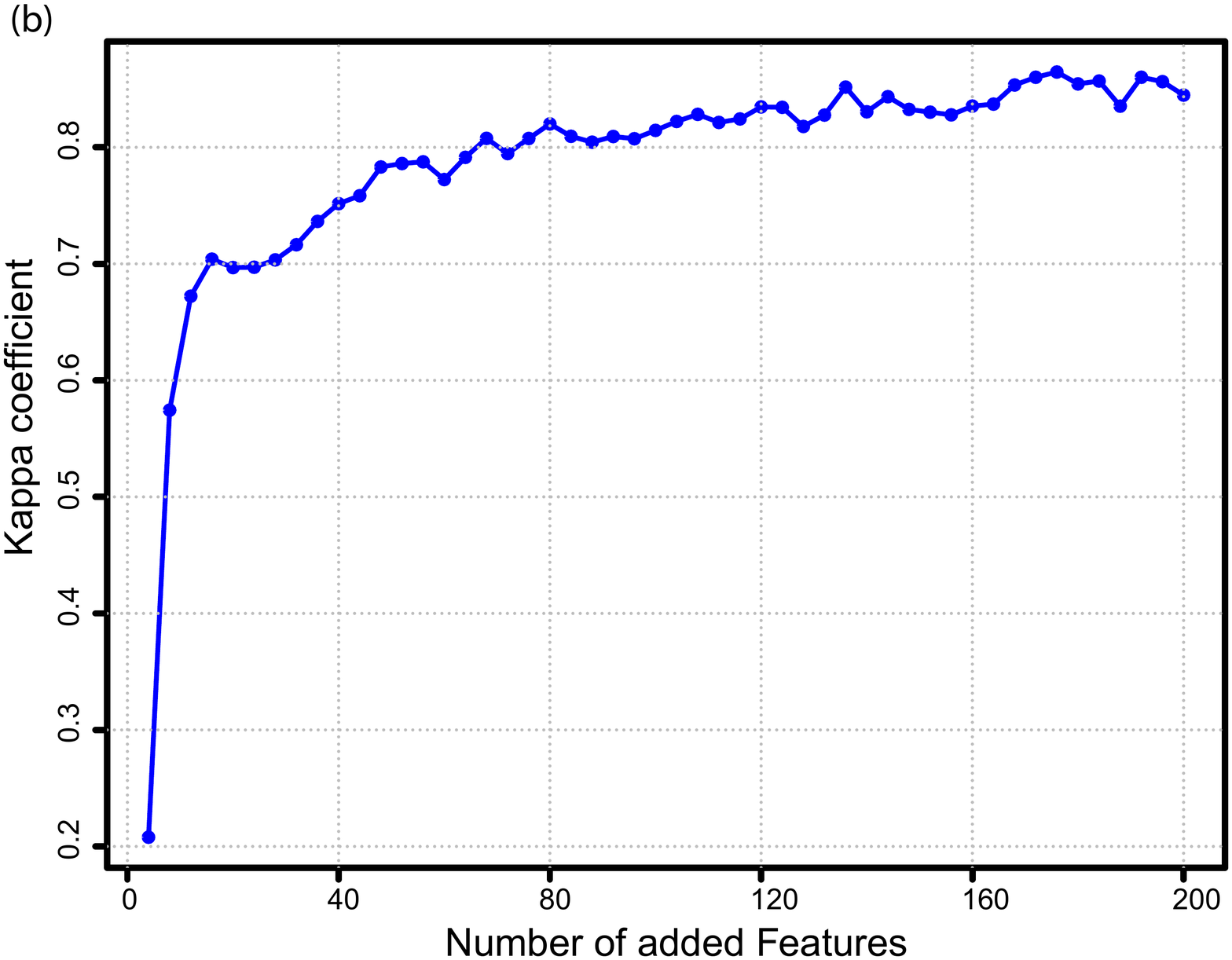}
\caption{Random forest results for each step of the UfsCov (a) the overall accuracy, (b) the Kappa coefficient. (Indian Pines dataset) }
\label{rfindian}
\end{figure}

\section{Conclusion}

The research introduced a space filling measure for the unsupervised feature selection problems. A new filter algorithm considered is based on the coverage measure. The proposed \textsf{UfsCov} algorithm minimises redundancy in data. The proposed algorithm showed its efficiency by testing on simulated and real world case studies including environmental data. Random forest results confirm the potential of space filling concept in the unsupervised feature selection problems. Finally, the \textsf{UfsCov} was programmed in R language and will be available on the CRAN repository in the \textsf{SFtools} library.

Future development could be in the adaptation of new measures based on space filling concept for machine learning use and data mining. Furthermore, it could be important to propose algorithms with a parallel CPU computing version and GPU computing to speed up the execution time.
\label{5}

\section{Acknowledgements}
\label{6}
This research was partly supported by the Swiss Government Excellence Scholarships for Foreign Scholars.

The authors would like to thank Nicola Deluigi for providing us with the Permafrost dataset. They also would like to thank Micheal Leuenberger, Jean Golay, and Fabian Guignard for fruitful discussions about machine learning.

\bibliography{xampl}
\bibliographystyle{elsarticle-num}

\end{document}